\documentclass[conference]{IEEEtran}
\IEEEoverridecommandlockouts
\usepackage{multirow}
\usepackage{tabularx}
\newcolumntype{M}[1]{>{\centering\arraybackslash}m{#1}}
\usepackage{amsmath}
\DeclareMathOperator*{\argmin}{arg\,min}
\usepackage{csquotes}
\usepackage{dsfont} % For indicator 1
\usepackage{subfig} %for subfloat
\usepackage{amssymb} % For \intercal
\usepackage{booktabs} % toprule
\usepackage{adjustbox} %adjustbox for table
\usepackage{bbold}      % Indicator function
\usepackage{comment}
\usepackage{hyperref} % The class enables hyperlinking by loading the package.
\usepackage[table]{xcolor} % rowcolor

\definecolor{purp}{rgb}{0.9,0.9,1.0}

\newtheorem{definition}{Definition}

\newcommand{\x}{\mathbf{x}}
\newcommand{\y}{\mathbf{y}}
\newcommand{\inputDim}{I}
\newcommand{\outputDim}{O}
\newcommand{\midDim}{W}
\newcommand{\reconError}{\boldsymbol{\rho}}
\newcommand{\predError}{\mathbf{e}}
\newcommand{\gaussCdf}{\Phi}
\newcommand{\gaussQuantile}{\gaussCdf^{-1}}
\newcommand{\Cov}{\boldsymbol{\Sigma}}
\newcommand{\Mean}{\boldsymbol{\mu}}
\newcommand{\empCDF}{\hat{F}}
\newcommand{\empQuantile}{\empCDF^{-1}}
\newcommand{\qhat}{\hat{q}}

\newcommand{\predInt}[1]{\epsilon_{\text{#1}}}
\newcommand{\score}[1]{s_{\text{#1}}\left(\x,\y\right)}
\newcommand{\ue}[1]{u(#1)}

\newcolumntype{P}[1]{>{\centering\arraybackslash}p{#1}}
\newcolumntype{Y}{>{\centering\arraybackslash}c}

\usepackage{cite}
\usepackage{amsmath,amssymb,amsfonts}
\usepackage{algorithmic}
\usepackage{graphicx}
\usepackage{textcomp}
\usepackage{xcolor}
\def\BibTeX{{\rm B\kern-.05em{\sc i\kern-.025em b}\kern-.08em
    T\kern-.1667em\lower.7ex\hbox{E}\kern-.125emX}}
\begin{document}

\title{Towards Trustworthy Vital Sign Forecasting: Leveraging Uncertainty for Prediction Intervals
% Towards Trustworthy Vital Sign Forecasting: Constructing Prediction Intervals from Uncertainty Estimates
% Previous Title: Uncertainty-to-Interval: Deriving Prediction Intervals from Uncertainty Estimates
% {\footnotesize \textsuperscript{*}Note: Sub-titles are not captured for https://ieeexplore.ieee.org  and should not be used
\thanks{This research was supported by the Ministry of Education, Singapore (Grant ID: RS15/23) and the College of Computing and Data Science, Nanyang Technological University.}
}

\author{Anonymous}

\author{\IEEEauthorblockN{1\textsuperscript{st} Li Rong Wang}
\IEEEauthorblockA{\textit{College of Computing and Data Science} \\
\textit{Nanyang Technological University}\\
\textit{\& A*STAR Centre for Frontier AI Research}\\
Singapore \\
0000-0003-3546-9354}
\and
\IEEEauthorblockN{2\textsuperscript{nd} Thomas C. Henderson}
\IEEEauthorblockA{\textit{School of Computing} \\
\textit{University of Utah}\\
Salt Lake City, United States\\
0000-0002-0792-3882}
\and
\IEEEauthorblockN{3\textsuperscript{rd} Yew Soon Ong}
\IEEEauthorblockA{\textit{College of Computing and Data Science} \\
\textit{Nanyang Technological University}\\
\textit{\& A*STAR Centre for Frontier AI Research}\\
Singapore \\
0000-0002-4480-169X}
\and
\IEEEauthorblockN{4\textsuperscript{th} Yih Yng Ng}
\IEEEauthorblockA{\textit{Swee Hock School of Public Health} \\
\textit{National University of Singapore}\\
Singapore \\
0000-0003-4598-1829}
\and
\IEEEauthorblockN{5\textsuperscript{th} Xiuyi Fan}
\IEEEauthorblockA{\textit{Lee Kong Chian School of Medicine} \\ \textit{\& College of Computing and Data Science} \\
\textit{Nanyang Technological University}\\
Singapore \\
0000-0003-1223-9986} 
}

\maketitle

\begin{abstract}
Vital signs, such as heart rate and blood pressure, are critical indicators of patient health and are widely used in clinical monitoring and decision-making. While deep learning models have shown promise in forecasting these signals, their deployment in healthcare remains limited in part because clinicians must be able to trust and interpret model outputs. Without reliable uncertainty quantification -- particularly calibrated prediction intervals (PIs) -- it is unclear whether a forecasted abnormality constitutes a meaningful warning or merely reflects model noise, hindering clinical decision-making. To address this, we present two methods for deriving PIs from the Reconstruction Uncertainty Estimate (RUE), an uncertainty measure well-suited to vital-sign forecasting due to its sensitivity to data shifts and support for label-free calibration. Our parametric approach assumes that prediction errors and uncertainty estimates follow a Gaussian copula distribution, enabling closed-form PI computation. Our non-parametric approach, based on k-nearest neighbours (KNN), empirically estimates the conditional error distribution using similar validation instances. We evaluate these methods on two large public datasets with minute- and hour-level sampling, representing high- and low-frequency health signals. Experiments demonstrate that the Gaussian copula method consistently outperforms conformal prediction baselines on low-frequency data, while the KNN approach performs best on high-frequency data. These results underscore the clinical promise of RUE-derived PIs for delivering interpretable, uncertainty-aware vital sign forecasts.
\end{abstract}


\section{Introduction}
Vital signs, such as heart rate, respiratory rate, and blood pressure, are fundamental clinical measurements that provide critical information about patients' health and body functions \cite{mok2015vital,cardona2015vital}. They are sensitive indicators of a patient's condition and are often the first signals of physiological distress \cite{brekke2019value,kenzaka2012importance}. Their abnormalities can precede visible symptoms, making them essential for early intervention \cite{eddahchouri2022effect,becking2023continuous}, especially in acute care, emergency medicine, intensive care, and hospital at home settings. 

Recently, machine learning methods have been explored to {\em predict} vital signs \cite{bhavani2022stress,shaik2024graph,changTransformerbasedDiffusionProbabilistic2024}. These approaches represent a shift from treating vital signs solely as inputs for risk score calculations to directly forecasting their future trajectories using patients’ electronic medical records. Despite promising results, almost all studies report only point-error metrics (e.g., MAE, RMSE), leaving clinicians unaware of the reliability of each forecast. Without calibrated {\em Prediction Intervals (PIs)}, it is impossible to determine whether a projected hypertensive episode constitutes a high-confidence warning or merely reflects noise in the model. Given the high-stakes nature of clinical decision-making, the absence of uncertainty quantification substantially limits the practical utility of these models.

To address this gap, we propose two PI estimation method based on uncertainty quantification, enabling vital sign prediction models to be better equipped for clinical deployment. Unlike traditional uncertainty quantification approaches such as Bayesian Neural Networks (BNNs) \cite{abdullahReviewBayesianDeep2022}, Monte Carlo dropout (MCD) \cite{gal2016dropout}, and deep ensembles \cite{lakshminarayananSimpleScalablePredictive2017}, which typically produce a single scalar \enquote{uncertainty score} for each prediction, our methods generate PIs that explicitly represent a range of plausible outcomes. The width of each interval intuitively reflects the model’s confidence in its forecast. For example, rather than outputting an abstract \enquote{uncertainty value of 5} for a heart rate prediction, our method would output a concrete interval such as 160 to 164 beats per minute.

For a patient $\x$, let a vital sign forecasting model produce a point prediction $\hat{\y}$, and let an {\em uncertainty estimation} model $\ue{\x}$ return a confidence measure. Given a target error tolerance level $\alpha \in (0,1)$, (typically $\alpha = 0.05$), we define a prediction interval by estimating the smallest $\epsilon$ such that
\begin{equation}\label{eqn:pi_condition}
    \Pr\left( |\y - \hat{\y}| \leq \epsilon \mid \ue{\x} \right) = 1 - \alpha,
\end{equation}
where $\y$ is the true value of the vital sign in the future. The resulting interval $[\hat{\y} - \epsilon, \hat{\y} + \epsilon]$ achieves the prescribed coverage level while adapting its width to the model's uncertainty $\ue{\x}$. 

A widely used method for deriving PIs is Conformal Prediction (CP). Conventional CP \cite{gammerman1998learning} sets the PI width $\epsilon$ to the appropriate quantile of absolute prediction errors (known as the non-conformity score) measured on a calibration set. Normalized CP \cite{papadopoulos2008normalized, renkemaConformalPredictionStochastic2024, khurjekarUncertaintyQuantificationDirectionofarrival2023, 
timansAdaptiveBoundingBox2024} extends this approach by normalizing the non-conformity score with a scalar uncertainty estimate, producing an uncertainty-conditioned PI. Building on this strong foundation, we enhance normalized CP's ability to model Equation~\ref{eqn:pi_condition} with two key improvements:
\begin{enumerate}
    \item {\bf Richer uncertainty-width mapping}. We allow more expressive nonlinear functions -- such as Gaussian copulas and K-nearest neighbours (KNN) —- beyond simple scalar multiplication to capture complex relationships between uncertainty estimates $\ue{\x}$ and interval half-width $\epsilon$.
    \item {\bf Multi-dimensional conditional.} We generalize CP to handle vector-valued     uncertainty estimates $\ue{\x} \in \mathbb{R}^K, K>1$, enabling tighter, context-aware intervals. 
\end{enumerate}

To achieve this, we employed the recently introduced multidimensional uncertainty measure, Reconstruction Uncertainty Estimate (RUE) \cite{rue_ias,korte2024confidence} for computing $\ue{\x}$. RUE has seen its success in time-series applications \cite{hao2024exploring,Li2025trading} and is designed for quantifying uncertainty from data shifts. RUE is particularly well suited to vital-sign forecasting because it supports label-free calibration, allowing continuous updates from unlabelled telemetry streams. Its reconstruction-error mechanism also remains sensitive to the covariate shifts common in bedside monitoring, such as sensor drift, changes in patient mix, and irregular sampling. These properties enable us to build reliable, context-aware prediction intervals without repeatedly retraining the base forecaster.

Leveraging on the multiple-dimensional outputs from RUE, we developed two strategies to derive PIs from RUE. The parametric approach, based on a Gaussian copula, assumes that the uncertainty scores and prediction errors follow a joint distribution, enabling closed-form computation of the conditional distribution in Equation \eqref{eqn:pi_condition}. The non-parametric approach, a k-nearest-neighbours (KNN) estimator, constructs this conditional distribution empirically by using prediction errors from validation points with similar reconstruction errors.

In vital sign prediction, a wide range of health signals are used, with sampling rates spanning from high frequencies (e.g., ECG at 1024 Hz and accelerometers at 50 Hz \cite{rahman2024exploring,shawen2020role}) to low frequencies measured in days (e.g., steps, heart rate, sleep status, and blood oxygen saturation \cite{ortiz2024data}). We evaluate our PI methods on two large public vital signs datasets sampled at minute and hour intervals to test their effectiveness across different frequencies. Our experiments demonstrate that the Gaussian-copula approach consistently surpasses CP baselines across several evaluation metrics—particularly on low-frequency data—whereas the KNN method achieves superior performance on high-frequency data.

%This paper makes the following contributions:
%\begin{enumerate}
%    \item It introduces two methods -- parametric and non-parametric -- for deriving PIs from RUE.
%    \item It presents a thorough comparison of PI performance between RUE-derived intervals and baseline uncertainty estimates with normalized CP in the context of ICU patient monitoring through time series prediction.
%\end{enumerate}

\section{Related Work}\label{sec:related}

\paragraph{Vital Sign Forecasting}
% Vital sign prediction
Recent work on vital sign forecasting spans a range of statistical and deep learning approaches. \cite{bhavani2022stress} employed both traditional time series models -- including AutoRegressive (AR), Moving Average (MA), and ARIMA -- as well as recurrent neural networks like LSTM and GRU to forecast pulse, oxygen saturation, and blood pressure. \cite{liu2019early} proposed a generative boosting framework that leverages a generative LSTM to first generate future sequences and then uses these synthetic steps to enhance the prediction of heart rate and systolic blood pressure by the predictive LSTM up to 20 minutes ahead. \cite{shaik2024graph} introduced a graph neural network architecture tuned via reinforcement learning and Bayesian optimisation to forecast heart rate up to one hour into the future. More recently, \cite{changTransformerbasedDiffusionProbabilistic2024} integrated Transformer-based sequence modelling with diffusion probabilistic methods to forecast heart rate, systolic, and diastolic blood pressure, demonstrating strong performance on physiological time series.

% Uncertainty + vital sign-ish 
While vital sign forecasting has been widely studied, few papers address the generation of prediction intervals. Notably, \cite{phetrittikun2021temporal} and \cite{he2025simultaneous} utilize the Temporal Fusion Transformer to forecast vital signs in ICU patients and generate prediction intervals via quantile regression. While quantile regression can be applied to any model architecture, it offers no formal coverage guarantees and tends to perform poorly in out-of-distribution settings. Despite these limitations, existing work has not explored alternative approaches -- such as deriving prediction intervals from distributional uncertainty estimates with CP -- for vital sign forecasting.

\paragraph{Uncertainty Estimation}
Uncertainty estimates measure how certain a model is about its prediction for each instance.
% Definition of uncertainty estimation
\begin{definition}[Uncertainty Estimation]
    Given an instance $\x\in \mathbb{R}^\inputDim$ and a trained prediction model $f$, an uncertainty estimation method $\sigma(\x;f): \mathbb{R}^\inputDim \rightarrow \mathbb{R}$ quantifies the model's uncertainty about its prediction on the instance.
\end{definition}

\begin{comment}
Each uncertainty estimation method quantifies at least one of the three types of uncertainty \cite{malininPredictiveUncertaintyEstimation2018a,durasovMasksemblesUncertaintyEstimation2021,hullermeier2021aleatoric}:
\begin{enumerate}
    \item \emph{Aleatoric Uncertainty}: Arises from noise in the data, e.g., due to sensor noise. % and is irreducible. 
    \item \emph{Epistemic Uncertainty}: Reflects uncertainty in model parameters and decreases with more training data, e.g., due to irregular sampling of patient vitals.
    \item \emph{Distributional Uncertainty}:  Arises from a data shift between the training set and the query instance, e.g., differing patient demographics between the training and prediction sets.
    % , at times categorized under epistemic uncertainty in the literature, 
\end{enumerate}
Our study focuses on quantifying distributional uncertainty.
\end{comment}

An established uncertainty estimation model is the Gaussian Process Regressor (GPR) \cite{williams2006gaussian}, which models data as a Gaussian process -- a distribution over functions -- and provides 
%epistemic 
uncertainty estimates through its covariance function. Recent developments include Sparse GPR \cite{titsias2009variational}, which improves computational efficiency.
% Additional Citations: wangIntuitiveTutorialGaussian2023,schulz2018tutorial
% However, traditional GPR suffers from high time complexity ($\mathcal{O}(n^3)$) and memory consumption ($\mathcal{O}(n^2)$) \cite{liu2020gaussian, wang2020intuitive}. Variations like Sparse GPR \cite{titsias2009variational} mitigate these issues by using a smaller set of learnable inducing points ($m \ll n$), reducing time complexity to $\mathcal{O}(nm^2)$.
%
% Another way to address GPR's scalability issues is by integrating uncertainty estimation into scalable models like neural networks. 
A more scalable approach, the
Bayesian Neural Network (BNN) \cite{jospinHandsOnBayesianNeural2022} models weights as probability distributions, reflecting 
%epistemic 
uncertainty. However, they are computationally expensive, difficult to train, and require significant architectural modifications, which can affect task performance \cite{yuDiscretizationInducedDirichletPosterior2024}.
%During inference, BNNs produce a distribution of outputs, with the standard deviation serving as an uncertainty estimate. However, BNNs are computationally expensive and difficult to train, even with techniques like Markov Chain Monte Carlo \cite{andrieu2013reversible} and variational inference \cite{blei2017variational}. 

Deep Ensemble and MCD approximate Bayesian inference in BNNs without altering neural architectures. Both estimate %epistemic 
uncertainty by sampling multiple predictions and computing their variance. In Deep Ensemble \cite{lakshminarayananSimpleScalablePredictive2017}, predictions are derived from individual models in the ensemble, while MCD \cite{gal2016dropout} generates predictions from a single model with dropout activated during inference. Deep Ensembles typically provide more reliable uncertainty estimates \cite{durasovMasksemblesUncertaintyEstimation2021} due to reduced prediction correlations but require longer training times. In contrast, MCD's multiple forward passes increase inference time, limiting its use in real-time applications.
% Techniques such as training multiple prediction heads \cite{linmansPredictiveUncertaintyEstimation2023a} or using model checkpoints with cyclical learning rates \cite{huangSnapshotEnsemblesTrain2017, garipovLossSurfacesMode2018} can reduce training costs but may increase prediction correlations, diminishing uncertainty estimate reliability.
%
Infer-Noise \cite{miTrainingFreeUncertaintyEstimation2022} (IN), like MCD, estimates 
%epistemic 
uncertainty by injecting Gaussian noise into intermediate layers during inference and computing the variance of predictions. However, it also requires multiple forward passes, increasing inference time.
Deep evidential learning (DEL) models \cite{ulmerPriorPosteriorNetworks,charpentierPosteriorNetworkUncertainty2020} are a class of single-pass uncertainty estimation methods. Instead of predicting the target directly, DEL models predict the parameters of a distribution, from which both a prediction and an uncertainty estimate are derived. For regression, \cite{aminiDeepEvidentialRegression2020} predicts the parameters of a Normal Inverse-Gamma distribution, using the expectation of variance 
%for aleatoric uncertainty 
and the variance of the mean for 
%epistemic 
uncertainty.
% the expected mean as the prediction and 
% Additional Citations: sensoyEvidentialDeepLearning2018a

\begin{figure*}[t]
    \centering
    % [trim={left bottom right top},clip]
    \includegraphics[width=\linewidth,trim={0 10.0838cm 0 0},clip]{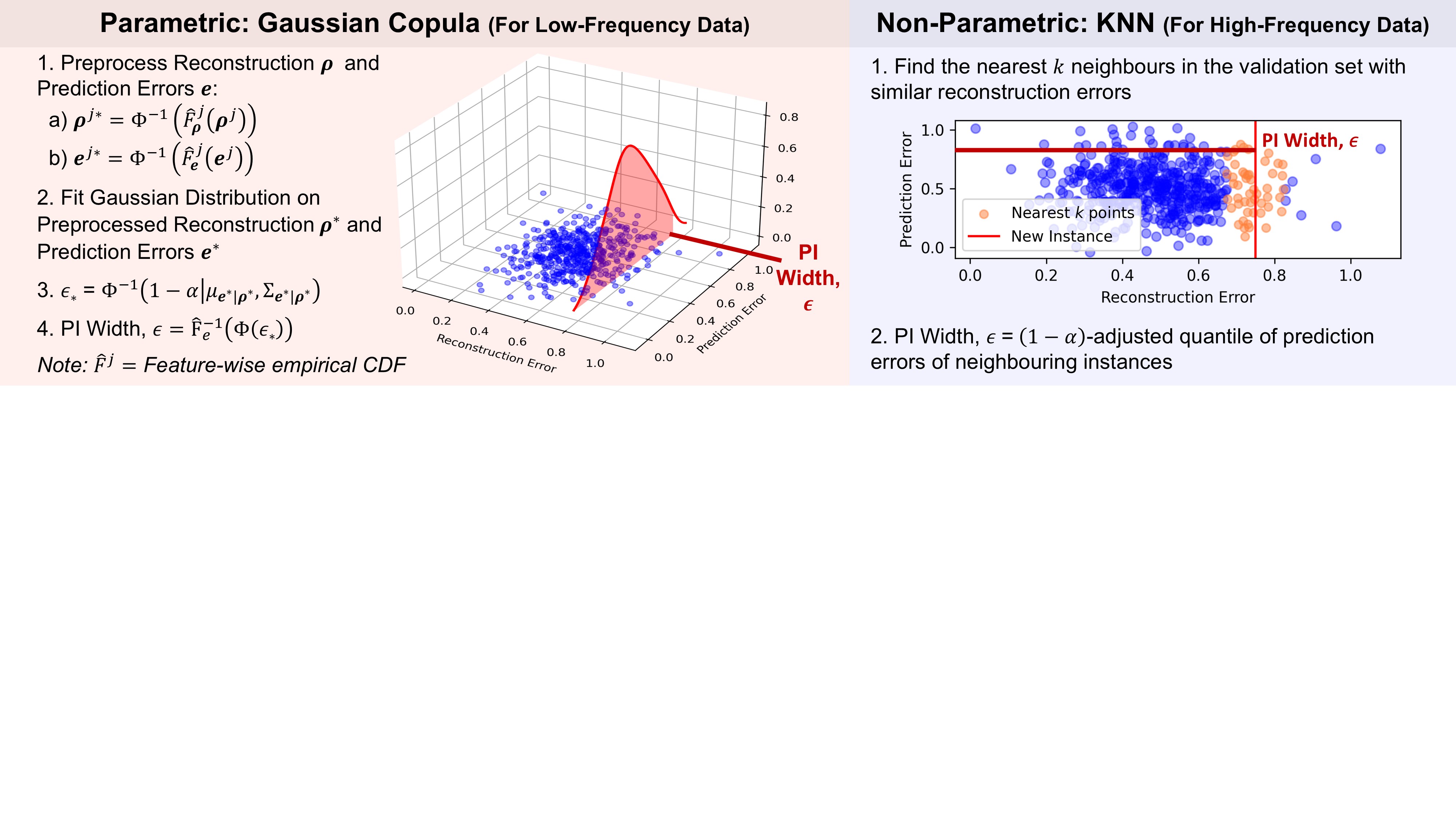}
    \caption{Summary of proposed prediction interval (PI) methods. For the parametric Gaussian Copula PI, reconstruction and prediction errors are first transformed into a Gaussian space using their empirical CDFs $\hat{F}$ and the inverse Gaussian CDF $\Phi^{-1}$. A joint Gaussian is then fitted to the transformed errors, and the PI width is computed from the $(1-\alpha)$-quantile of the conditional distribution of prediction errors given reconstruction errors, followed by mapping back to the original space using the inverse transforms. For the KNN PI, the PI width is computed as the $(1-\alpha)$-adjusted quantile of the prediction errors of the instance's $k$ nearest neighbors with similar reconstruction errors.}
    \label{fig:pi_intervals}
    %\vspace{-5pt}
\end{figure*}

Deterministic Uncertainty Methods (DUMs) \cite{postelsPracticalityDeterministicEpistemic2022a,charpentierTrainingArchitecturePrior2023}, including RUE, differ from DEL models as they do not modify the training objective. DUMs estimate uncertainty based on an instance's distance to the training set in the model's latent space, measured relative to neighboring instances of different classes \cite{mandelbaumDistancebasedConfidenceScore2017} or class centroids \cite{amersfoortUncertaintyEstimationUsing2020,apostolopoulouRateDistortionViewUncertainty2024}. RUE estimates the distance to the training set using reconstruction error; instances dissimilar to the training set are poorly reconstructed, resulting in higher errors.
% Additional Citations: winkensContrastiveTrainingImproved2020,postelsHiddenUncertaintyNeural2021,mukhotiDeepDeterministicUncertainty2022

\paragraph{Conformal Prediction (CP)}
CP \cite{gammerman1998learning} generates PIs with coverage $1-\alpha$ from any prediction model. This paper focuses on split CP \cite{papadopoulos2007conformal}, which uses a fixed calibration set for computational efficiency. Split CP involves three main steps:
\begin{enumerate}
    \item Computes a conformal score $\score{}$ for each instance in the calibration set, typically quantifying the model’s error (e.g.,  $\score{}=\left|\y - \hat{\y}\right|$ ; $\y$ and $\hat{\y}$ refer to the true and predicted targets, respectively). 
    \item Calculates $\hat{q}$, the $1 - \alpha$ adjusted quantile of conformal scores within the calibration set.
    \item PI for any new instance is derived as $[\hat{\y} - \hat{q},~\hat{\y} + \hat{q}]$. 
\end{enumerate}
% Additional citations: tibshirani2023conformal,dewolfValidPredictionIntervals2023
However, this approach results in a PI that is uniform in width, regardless of the specific input. The state-of-the-art approach, normalized CP, generates PIs conditioned on one-dimensional uncertainty estimates, $\ue{\x}$ \cite{papadopoulos2008normalized,angelopoulosGentleIntroductionConformal2022}. Normalized CP achieves this by modifying the score function $\score{}$; rather than using only prediction errors, normalized CP divides the errors by uncertainty: 
% Additional citations: khurjekarUncertaintyQuantificationDirectionofarrival2023
\begin{equation} 
\score{} = \frac{\left|\y-\hat{\y}\right|}{\ue{\x}} \label{eqn:norm_cp_score} 
\end{equation} 
Normalized CP derives the PI as $[\hat{y}-\ue{\x}\hat{q},~ \hat{y}+\ue{\x}~\hat{q}]$, conditioning the PI on the uncertainty of each instance. Note that normalized CP allows one to use any uncertainty estimate to form the PIs (e.g., conformal RUE). These methods differ fundamentally from our approaches, which compute the PIs directly without using Equation~\ref{eqn:norm_cp_score}. 

%\paragraph{Multi-dimensional CP}
Multi-dimensional conformal prediction methods aim to compute PIs based on multi-dimensional conformity scores. Most approaches reduce these multi-dimensional scores to a single dimension. For example, \cite{luo2024weighted} uses a weighted sum of conformity scores. \cite{klein2025multivariate} applies optimal transport to map conformity score vectors to a uniform ball distribution, using the norm as a one-dimensional conformity score. \cite{xu2024conformal} employs the Mahalanobis distance of the multi-dimensional conformity score as a scalar conformity measure. \cite{tawachi2025multi} clusters the conformity score space and computes scalar conformity scores within each cluster. However, these methods produce PIs with uniform width across inputs and are not input-adaptive, unlike our proposed method. Only \cite{lee2025flow} addresses this limitation by using conditional normalizing flows to map multi-dimensional conformity scores into a Gaussian latent space, where the norm serves as a one-dimensional conformity score. 
% During inference, the flow maps the coverage region in latent space back to the original score space to generate input-dependent PIs. Importantly, because the flow is conditioned on the model’s context representation, the resulting PIs vary with the input. 
Nevertheless, none of these methods leverage uncertainty estimates in PI construction nor apply their approaches to healthcare datasets, as is done in our proposed methods.

\section{Methodology}\label{sec:method}
% In this section, we provide an overview of the proposed methods for prediction interval generation from Reconstruction Uncertainty Estimates.
\paragraph{Reconstruction Uncertainty Estimate}
\cite{rue_ias,korte2024confidence} introduces RUE, a distributional uncertainty estimate calculated in a single pass without modifying model architecture or training objective.
% inspired by anomaly detection with autoencoders. 

Consider a model $f_{\phi, \psi}$, trained with input $\x \in \mathbb{R}^\inputDim$ and output $\y \in \mathbb{R}^\outputDim$. The model is parameterized by $\phi$ and $\psi$, which denote the feature extractor $f_{\phi}$ and prediction head $f_{\psi}$. In a regression problem, $f_{\phi}$ outputs a feature vector of size $k$, which is used by $f_{\psi}$ to compute continuous output values $\hat{\y}$, i.e., $f_{\phi,\psi} : \mathbb{R}^\inputDim \mapsto \mathbb{R}^\outputDim$. The model $f_{\phi, \psi}$ is typically trained using gradient descent to minimize the discrepancy between the true $\y$ and predicted $\hat{\y}=(f_\psi \circ f_\phi)(\x)$ target:
\begin{equation}
\label{eqn:prediction}
f_{\phi, \psi} = \argmin_{\phi, \psi} ||\y - (f_\psi \circ f_\phi)(\x)||^2.
\end{equation}
To compute RUE, another model — referred to as the decoder and denoted as $g: \mathbb{R}^\midDim \mapsto \mathbb{R}^\inputDim$ — is trained to reconstruct the input $\x$ of the prediction model from the latent representation produced by its feature extractor $f_{\phi}$, which has $W$ elements. $g$ is trained using gradient descent to minimize the discrepancy between $\x$ and $\hat{\x} = (g \circ f_{\phi})(\x)$:
\begin{equation}
\label{eqn:decoder}
  g = \argmin_g ||\x - (g \circ f_{\phi})(\x)||^2.
\end{equation}
$f_\phi$ and $g$ form an autoencoder with one distinction: instead of training the encoder and decoder simultaneously, RUE adopts a two-step process to avoid compromising the prediction model's performance. First, the encoder $f_\phi$ is trained as part of the prediction model (Equation \ref{eqn:prediction}). Once $f_\phi$ is fully trained, the decoder $g$ is trained separately (Equation \ref{eqn:decoder}).

\begin{definition}[Reconstruction Uncertainty Estimate]
    The reconstruction uncertainty estimate (RUE) of instance $\x$ is the difference between the actual and reconstructed input \footnote{$|v|_1$ is the L1-norm of vector $v$.}:
    \begin{align*}
        \sigma(\x;f) := \left|\mathbf{\x} - (g \circ f_\phi)(\mathbf{\x})\right|_1.
    \end{align*}
\end{definition}

% for an instance $\mathbf{\x} \in \mathbb{R}^\inputDim$ and its corresponding label $\mathbf{\y} \in \mathbb{R}^\outputDim$, 
RUE is hypothesized to reliably estimate prediction error, showing a positive correlation with prediction error:
\begin{equation}
\label{eqn:prop}
  \left|\mathbf{\y} - f_{\phi, \psi}(\mathbf{\x})\right|_1 \leftrightarrow \sigma(\x;f).
\end{equation}
The above hypothesis is grounded in two assumptions:
\begin{enumerate}
  \item The model $f_{\phi,\psi}$ performs well on inputs similar to the training set but poorly on unfamiliar inputs.
  \item The decoder $g$, trained on the same set, reconstructs familiar inputs well, yielding low reconstruction errors (small $\sigma$), but struggles with unfamiliar inputs, resulting in high errors (large $\sigma$).
\end{enumerate}

\paragraph{RUE-derived Prediction Intervals}

We have devised two methods to estimate PIs from feature-wise reconstruction $\reconError$ and prediction errors $\predError$ (Figure~\ref{fig:pi_intervals}). 
\begin{definition}[Feature-wise Reconstruction Error]
Given an instance $\x = [x^1, \ldots, x^\inputDim]$ and its reconstruction $\hat{\x} = [\hat{x}^1, \dots, \hat{x}^\inputDim]$, its feature-wise reconstruction error is:
\begin{align*}
    \reconError = \left[\left|x^1-\hat{x}^1\right|, \ldots, \left|x^\inputDim-\hat{x}^\inputDim\right|\right]
\end{align*}
\end{definition}
\begin{definition}[Output-wise Prediction Error]
Given an instance $\x$, its target $\y = [y^1, \dots, y^\outputDim]$ and the prediction $\hat{\y} = [\hat{y}^1, \dots, \hat{y}^\outputDim]$, its output-wise prediction error is:
\begin{align*}
    \predError = \left[\left|y^1-\hat{y}^1\right|, \ldots, \left|y^\outputDim-\hat{y}^\outputDim\right|\right]
\end{align*}
\end{definition}
Assuming a symmetric PI around the predicted target $\hat{\y}$:
\begin{equation}\label{eqn:symmetrical_pi}
L_i = \hat{\y}-\predInt{},~ U_i=\hat{\y}+\predInt{}
\end{equation}
We can modify Equation~\ref{eqn:pi_condition}, simplifying the PI generation problem to the task of estimating the conditional distribution of the prediction error $\predError$ given the reconstruction error $\reconError$.
\begin{align}
    \Pr \left(\predError_i \leq \predInt{} \mid \reconError_i \right) = 1-\alpha \quad \equiv \quad
    F_{\predError_i\mid\reconError_i}(\predInt{})=1-\alpha \label{eqn:updated_pi_condition}
\end{align}
Hence, each of the proposed methods aims to estimate the conditional distribution of prediction error given reconstruction error (Equation~\ref{eqn:updated_pi_condition}) and can be viewed as an extension of CP conditioned on feature-wise reconstruction error.

\paragraph{Gaussian Copula PI}
% The conditional Gaussian PI assumes that both reconstruction errors $\reconError$ and prediction errors $\predError$ follow a multivariate Gaussian distribution. However, this assumption may not hold for all datasets. We could instead 
The Gaussian Copula PI models the marginal distribution of each error using its empirical distribution and estimates the conditional distribution in Equation~\ref{eqn:updated_pi_condition} with a Gaussian copula \cite{aasExplainingIndividualPredictions2021}. Using a Gaussian copula is motivated by the tendency of low-frequency health signals to approximate a Gaussian distribution, making the copula well-suited to model their dependencies effectively.
% capture dependencies between variables with a Gaussian copula \cite{aasExplainingIndividualPredictions2021}.

In practice, the Gaussian Copula PI can be interpreted as modelling the reconstruction errors $\reconError$ and prediction errors $\predError$ using a Gaussian distribution, with additional pre- and post-processing steps. The reconstruction $\reconError$ and prediction errors $\predError$ are preprocessed using output-wise empirical CDFs $\empCDF$ and the inverse CDF of a Gaussian distribution, $\gaussQuantile$:
\begin{align}
   \reconError^{j*} = \gaussQuantile \left(\empCDF_{\reconError}^j\left(\reconError^j\right)\right),~
   \predError^{j*} = \gaussQuantile \left(\empCDF_{\predError}^j\left(\predError^j\right)\right)
\label{eqn:gauss_copula}
\end{align}

After preprocessing, we fit a multivariate Gaussian distribution to the preprocessed reconstruction $\reconError^*$ and prediction errors $\predError^*$ and compute the marginal PI as the $(1-\alpha)$-quantile of the conditional Gaussian distribution of preprocessed prediction errors $\predError^*$ given preprocessed reconstruction errors $\reconError^*$.
% following the same procedures as conditional Gaussian PI. 
\begin{equation}
    \predInt{*} = \gaussQuantile
    \left(1-\alpha \mid 
        \Mean_{\predError^* \mid \reconError^*}, 
        \Cov_{\predError^* \mid \reconError^*}
    \right)
\end{equation}

To derive the PI from the marginal PI, $\predInt{*}$, we apply the following post-processing step:
\begin{equation}
    \predInt{copula} =  \empQuantile_{\predError}\left(\gaussCdf\left(\predInt{*}\right)\right)
    \label{eqn:pi_copula}
\end{equation}
$\empQuantile_{\predError}$ represents the empirical quantile function for prediction errors $\predError$, and $\gaussCdf$ is the CDF of a Gaussian distribution.

Compared to CP, the Gaussian copula PI replaces the discrete score function of CP with the CDF of a Gaussian distribution for the preprocessed prediction errors $\predError^*$ conditioned on the preprocessed reconstruction errors $\reconError^*$:
\begin{equation}
    \score{copula} = \gaussCdf(\predError^* \mid \Mean_{\predError^* \mid \reconError^*}, \Cov_{\predError^* \mid \reconError^*})
\label{eqn:score_copula}
\end{equation}
Additionally, instead of applying a multiplicative correction factor to $\qhat$, we transform $\qhat$ using the CDF of a Gaussian and the inverse of the empirical CDF $\empQuantile_{\predError}$ (Equation~\ref{eqn:pi_copula}). 

\paragraph{K-Nearest Neighbours PI}\label{sec:knn}
The K-Nearest Neighbors (KNN) PI is a non-parametric method that estimates the conditional distribution in Equation~\ref{eqn:updated_pi_condition} empirically. KNN is ideal for high-frequency signals because it can capture complex local patterns in the data without assuming a specific global distribution, allowing flexible modeling of rapid physiological variations. It computes the PI for a given instance, $\mathbf{x}$, by using the prediction errors, $\predError$, of its neighbours in the reconstruction error $\reconError$-space. This is achieved through the following procedure:
\begin{enumerate}
    \item Find the nearest $k =\lfloor \sqrt{n_v} \rceil$ neighbours in the validation set (of size $n_v$) with similar $\reconError$ to $\mathbf{x}$.\label{itm:knn}
    \item Compute the prediction errors $\predError$ of the $k$ neighbours.\label{itm:abs_error} 
    \item The PI of $\x$ is the $(1-\alpha)$-adjusted quantile ($\lceil (k+1)*(1-\alpha) \rceil/k$) of prediction errors $\predError$ of its neighbours. \label{itm:percentile} 
\end{enumerate}

\begin{figure*}[htbp]
    \centering
    % Answer: [trim={left bottom right top},clip]
    \subfloat[][High-Frequency Data: MIMIC]{
        \includegraphics[width=0.5\linewidth]{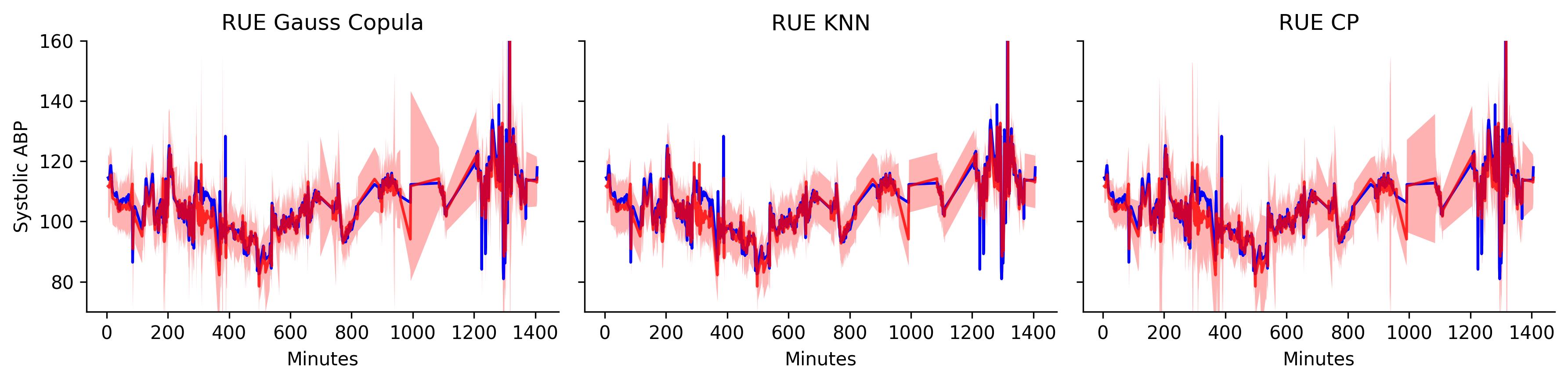}
        \label{fig:pi_lineplot_comparison_mimic} 
    }
    \subfloat[][Low-Frequency Data: PhysioNet]{
        \includegraphics[width=0.5\linewidth]{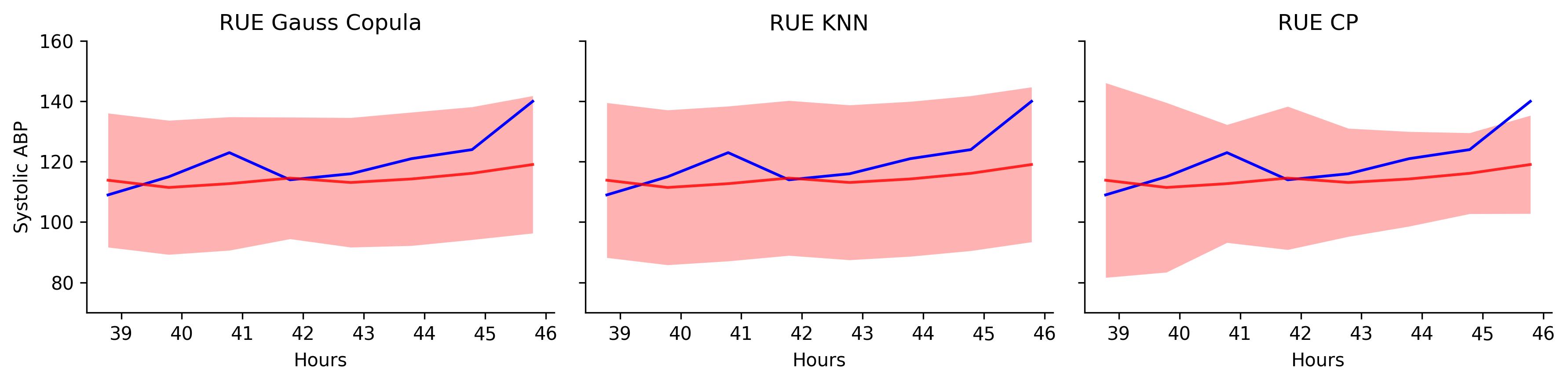}
    \label{fig:pi_lineplot_comparison_physionet}
    }
    \caption{Line plots showing the prediction intervals (light red shading), model predictions (dark red line), and true signal (blue line) for each prediction interval method at time horizon $t+1$. For MIMIC, we present the intervals for patient \enquote{465n} on the \enquote{ABPsys (mmHg)} feature. For PhysioNet, we show the intervals for patient \enquote{142692} on the \enquote{SysABP} feature.}%
    \label{fig:pi_lineplot_comparison}%
\end{figure*}

The KNN PI method estimates the distribution of PIs conditioned on the reconstruction error through Steps \ref{itm:knn} and \ref{itm:abs_error}. Step \ref{itm:percentile} calculates the PI with $1-\alpha$ coverage from this estimated distribution. For smaller validation sets ($n_v < \left(\frac{2}{\alpha} - 1\right)^2$), using the rule-of-thumb setting $k = \lfloor \sqrt{n_v} \rceil$ may lead to poor PI generation, as the adjusted quantile causes the interval to rely on the maximum neighboring prediction errors -- making it highly susceptible to outliers in the validation set. Instead, choosing a $k$ greater than or equal to $(\frac{2}{\alpha} - 1)$ (e.g., 80 in our experiments on PhysioNet) yields better PIs. We include ablation studies in the code repository, varying the KNN parameter $k$ to assess its impact on prediction interval performance.

% Additionally, using a percentile instead of a maximum also reduces the impact of outliers in the validation set on the PI.

Viewed through the lens of CP, the KNN PI uses the same score function as standard CP but limits the calibration set to instances with similar reconstruction errors, conditioning the PI on uncertainty.

\section{Experiment}\label{sec:experiment}
% In this section, we outline the experimental setup, including the datasets, evaluation metrics, and baseline models used. We then present the results comparing RUE and RUE-derived prediction intervals with existing methods.

\subsection{Data Sets}
To evaluate the effectiveness of RUE-derived PI methods for vital-sign prediction models, we applied them to two large, publicly available datasets: MIMIC and PhysioNet. MIMIC contains minute-level vital signs capturing detailed physiological changes, while PhysioNet provides hour-level data that reflects longer-term trends. This difference in temporal resolution (Figure~\ref{fig:pi_lineplot_comparison}) allows us to test PI methods across varying signal dynamics. Additionally, the datasets differ in calibration set size -- PhysioNet has a smaller calibration set -- enabling us to examine the impact of limited calibration data on PI performance.

\subsubsection{\textbf{MIMIC}}
The MIMIC dataset \cite{moody1996database} comprises vital signs from 121 intensive care unit (ICU) patients, each with a potentially different set of signals. After preprocessing (see Appendix~\ref{sec:preprocessing}), the dataset consists of 59,566 instances, with 38,769, 8,765, and 12,032 instances in the training, validation, and test sets, respectively. With the preprocessed data, we trained the prediction model, denoted as $f_{\phi, \psi}(\mathbf{\x})$, to forecast all six patient's states for one, two and three minutes into the future ($y_{t+1}$, $y_{t+2}$, $y_{t+3}$), utilizing the patient's signals from the past 5 minutes ($x_{t-5}$ to $x_{t}$). 

\subsubsection{\textbf{PhysioNet}}
The PhysioNet Challenge 2012 dataset \cite{silva2012predicting} comprises signals from 2,485 intensive care unit (ICU) patients. We selected 5 vital signs from the 37 available, as they had the fewest missing values: Diastolic ABP (Arterial Blood Pressure) (mmHg), Mean ABP (mmHg), Systolic ABP (mmHg), Heart Rate (bpm), and Urine Output (mL). Set A and Set B were used for training and testing, respectively, following the Challenge's specifications. After preprocessing, the dataset consisted of 25,105 instances, with 11,608, 1,244, and 12,253 instances in the training, validation, and test sets, respectively. We trained the prediction model to forecast all five patient's states for one, two and three hours into the future ($y_{t+1}$, $y_{t+2}$, $y_{t+3}$), utilizing the patient's signals from the past 5 hours ($x_{t-5}$ to $x_{t}$). 

\subsection{Baseline Models}
\begin{table*}[htbp]
    \setlength{\tabcolsep}{4pt}
    \renewcommand{\arraystretch}{1.6}
    \centering
    \caption{Brief descriptions of each prediction interval method, the calibration set and conformal score used, how the prediction width is computed, and whether the method is parametric or non-parametric. \enquote{1-D UE} indicates whether the method can be conditioned on one-dimensional uncertainty estimates, while \enquote{Multi-D UE} indicates whether the method can be conditioned on multi-dimensional uncertainty estimates. We indicate our proposed methods with \underline{underlining}.}
    \scriptsize
    \begin{tabular}{|P{1.6cm}||P{4cm}|P{1.2cm}|P{2.3cm}|P{2.5cm}|Y|P{0.7cm}|P{1cm}|}
    \hline
        \textbf{Prediction Interval} & \textbf{Description} & \textbf{Calibration Set} & \textbf{Conformal Score} & \textbf{Prediction Width} & \textbf{Parametric} & \textbf{1-D UE} & \textbf{Multi-D UE} \\ \hline
        \underline{Gauss Copula} &  Fit Gaussian Copula on uncertainty scores and prediction errors. & Whole & $\gaussCdf(\predError^* \mid \Mean_{\predError^* \mid \reconError^*}, \Cov_{\predError^* \mid \reconError^*})$ & $\empQuantile_{\predError}\left(\gaussCdf\left(\hat{q}\right)\right)$ & Yes & Yes & Yes \\ 
    \cline{1-2} \cline{4-8}
    \hline
        \underline{KNN} & CP with a calibration set of instances with similar uncertainty. &  $k$ instances with similar uncertainty & \multirow{4}{2cm}{\centering $\left|\y-\hat{\y}\right|$}  & $\hat{q}_a=$~$(1-\alpha)$-th adjusted quantile of scores & No & Yes & Yes  \\ 
    \cline{1-3} \cline{5-8}
        Conformal Prediction (CP) & PI widths are estimated as the adjusted quantile of calibration set scores. & \multirow{3}{*}{Whole} & ~ & $\hat{q}_a$ & No & No & No \\ 
    \cline{1-2} \cline{4-8}
        Normalised CP & Extends CP by normalizing the conformal score with uncertainty. & & $\frac{\left|\y-\hat{\y}\right|}{\ue{\x}}$  & $\hat{q}_a\times \ue{\x}$ & No & Yes & No \\ \hline
    \end{tabular}
    \label{tab:pi_related_work_comparison}
\end{table*}

In our experiments, we set $\alpha=0.05$ and compare our proposed PIs against the following baselines:

\paragraph{\textbf{RUE CP}} We compare the multidimensional RUE methods with aggregated single-dimension RUE combined with normalized conformal prediction (CP) \cite{papadopoulos2008normalized}, to demonstrate the benefits of conditioning on multidimensional RUE.

Additionally, we compare our PIs with those derived by applying normalized CP \cite{papadopoulos2008normalized} to five other uncertainty estimates. We selected single-model uncertainty estimates -- comparable to RUE -- that capture different types of uncertainty.

\paragraph{\textbf{Monte Carlo Dropout (MCD) CP}} We applied a 0.20 dropout rate to the prediction model's penultimate layer and generated 10 predictions per input, following hyperparameters used in \cite{gal2016dropout}.

\paragraph{\textbf{Sparse Gaussian Process Regressor (SGPR) CP}} We used \textit{GPFlow}’s implementation of SGPR \cite{GPflow2017,titsias2009variational} with a radial basis function kernel, selecting 39 and 103 random inducing variables (0.1\% and 1\% of training data) for MIMIC and PhysioNet respectively. 

\paragraph{\textbf{Infer-Noise (IN) CP}} We select $\sigma \in [0.00001, 0.5]$ \cite{miTrainingFreeUncertaintyEstimation2022} via grid search to minimize mean absolute prediction error and sample 10 predictions per input during inference.

\paragraph{\textbf{Bayesian Neural Network (BNN) CP}}: We implemented BNNs \cite{jospinHandsOnBayesianNeural2022} using \textit{torchbnn} \cite{lee2022graddiv} with Bayesian linear layers (prior mean: 0, standard deviation: 0.1) and sampled 10 predictions per input.
% All BNN models were trained using the Adam optimizer with default Keras settings. We selected the number and width of hidden layers based on the configuration that minimized mean squared error (for both: $d = {2, 3}$, $w = {32, 64, 128, 256}$). 

\paragraph{\textbf{Deep Evidential Regression (DER) CP}}: We implemented DER using \textit{TorchUncertainty} \cite{torchUncertainty} with regularization weight = 0.01, following \cite{aminiDeepEvidentialRegression2020}.
% We selected the number and width of hidden layers ($d = {2, 3, 4}$, $w = {128, 256, 512}$) to minimize mean squared error.

\subsection{Evaluation Metrics} \label{sec:experiment_metrics}
We assess PI quality using three metrics derived from the PI cost function in \cite{kabir2020neural}:
\paragraph{\textbf{Coverage Penalty (CovP)}}
CovP measures the deviation of Prediction Interval Coverage Probability (PICP) from the ideal $1-\alpha$ coverage, with lower CovP indicating a better PI.  PICP is the proportion of points within the PI.
\begin{equation}\label{eqn:metrics_cp}
    \text{CovP} = (1-\alpha+\delta-\text{PICP})^2,~\text{PICP} = \frac{1}{n}\sum_{i=1}^n \mathds{1}_{y_i\in[L_i, U_i]}
\end{equation}
$n$ represents the total number of points, $\y_i$ denotes the $i$th ground-truth output, $L_i$ and $U_i$ stand for the lower and upper bounds of the PI and $\delta=\alpha/50$ is the coverage margin.

\paragraph{\textbf{Prediction Interval Normalized Average Width (PINAW)}} The average normalized size of the PI. When two PIs have the same CP, the PI with the smaller PINAW is better.
\begin{equation}\label{eqn:pinaw}
    \text{PINAW} = \frac{1}{n\times R}\sum_{i=1}^n (U_i- L_i)
\end{equation}
$R$ is the range of the output variable.

\paragraph{\textbf{Coverage Width Failure Distance Criteria (CWFDC)}} It combines the two metrics above with Prediction Interval Normalized Average Failure Distance (PINAFD) to quantify the overall quality of the prediction interval (PI); a lower CWFDC indicates a better PI.
\begin{equation}
    \text{CWFDC} = \text{PINAW} + \rho\cdot\text{PINAFD} + \beta\cdot\text{CovP}
\end{equation}
Here, we set $\rho = 1$ and $\beta = 1000$, following the parameter values specified in \cite{kabir2020neural}.

PINAFD is the average normalized distance of points outside the PI. A smaller PINAFD is preferable, as it indicates that outliers are closer to the PI.
\begin{equation}\label{eqn:pinafd} 
    \text{PINAFD} = \frac{
    \sum_{i=1}^n \mathds{1}_{y_i\notin[L_i, U_i]} \min(\left|y_i-U_i\right|, \left|L_i-y_i\right|)}{
    R\times \sum_{i=1}^n (\mathds{1}_{y_i\notin[L_i, U_i]}) 
    }
\end{equation}
PINAFD is zero in the absence of outliers.

\subsection{Experiment Results}\label{sec:results}
\begin{figure*}[ht]
    \centering
    % Answer: [trim={left bottom right top},clip]
    \subfloat[][MIMIC]{
        \includegraphics[width=\linewidth]{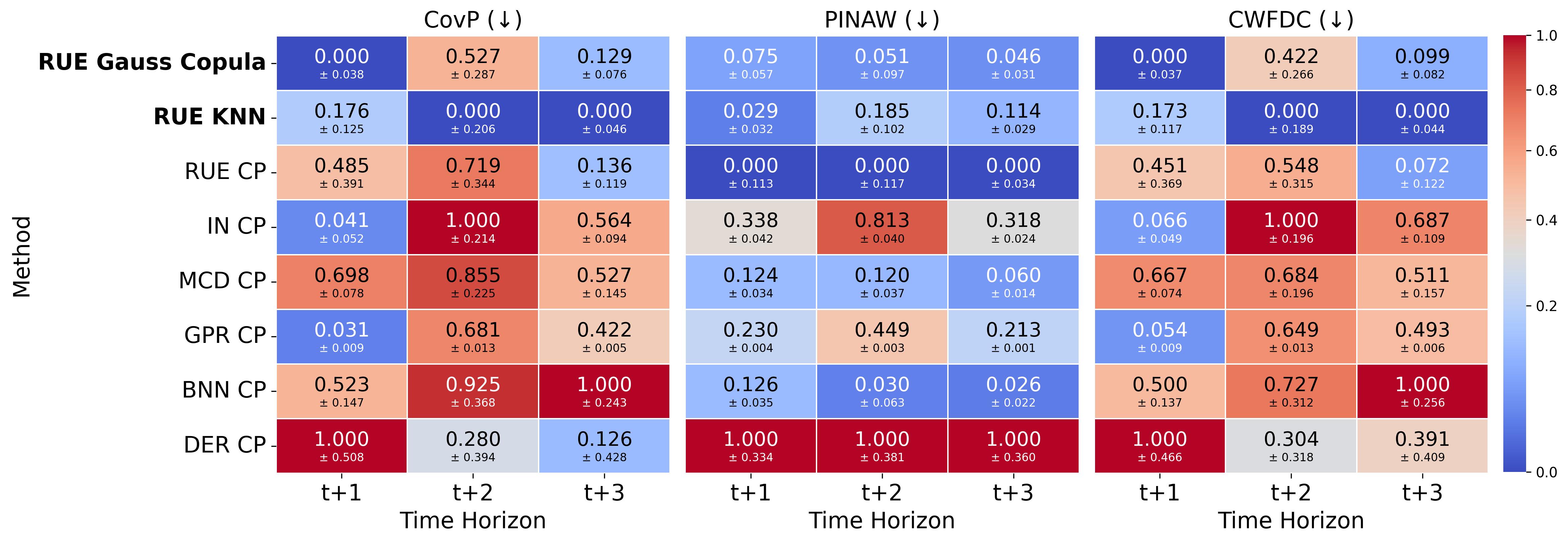}
        \label{fig:pi_comparison_mimic}
    }\\
    \subfloat[][PhysioNet]{
        \includegraphics[width=\linewidth]{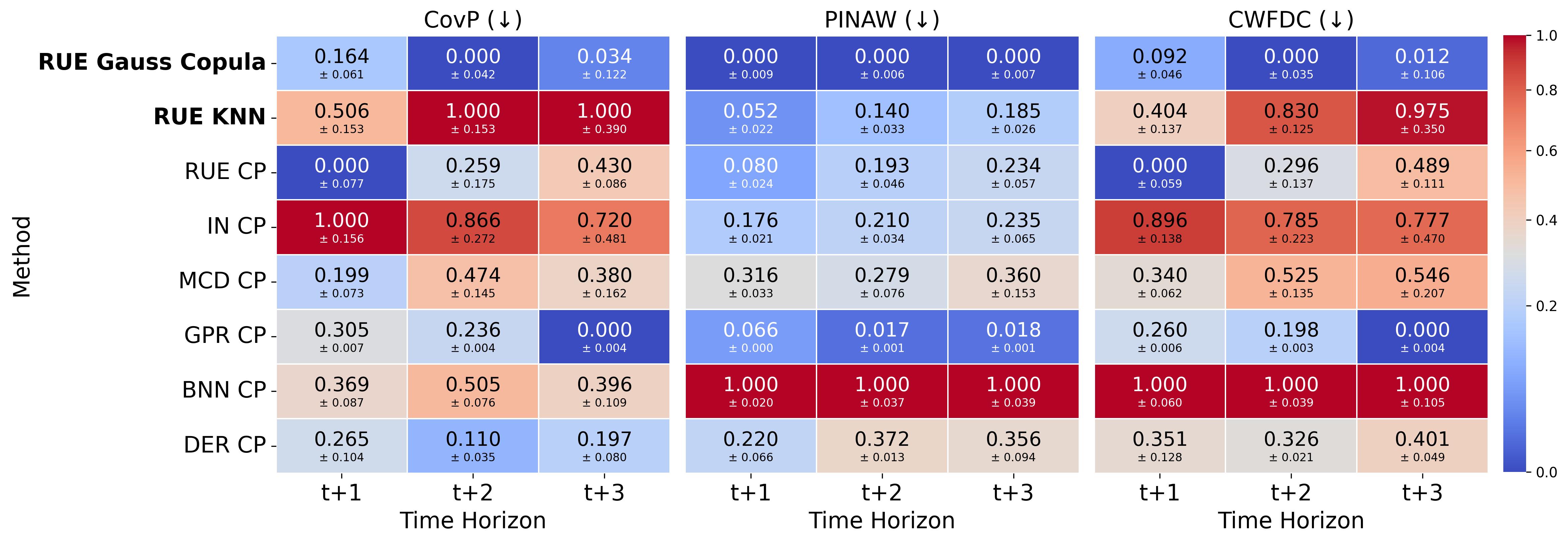}
        \label{fig:pi_comparison_snp}
    }
    \caption{Heatmaps comparing prediction intervals in terms of Coverage Penalty (CovP; $\downarrow$), Prediction Interval Normalized Average Width (PINAW; $\downarrow$) and Coverage Width Failure Distance Criteria (CWFDC; $\downarrow$) on MIMIC and PhysioNet. For easier analysis, we min-max normalise each metric within each time horizon. The proposed PIs are highlighted in \textbf{bold text}. CovP measures the deviation of the proportion of points within the PI from the ideal coverage of 0.95, while PINAW measures PI size. CWFDC measures the overall quality of the PIs. Since all three metrics are penalties, lower values (in \textcolor{blue}{blue}) are preferable. Each row represents a PI's performance; an ideal row would be entirely deep blue. However, trade-offs between the metrics make this unlikely. Comparing CWFDCs, the Gaussian Copula PI is the best PI, achieving the lowest CovP across datasets without compromising PINAW.}%
    \label{fig:pi_comparison}%
    %\vspace{-10pt}
\end{figure*}

Figure~\ref{fig:pi_comparison} compares all PIs on the MIMIC and PhysioNet datasets in terms of coverage penalty (CovP), PI width (PINAW), and overall PI quality (CWFDC).
\paragraph{\textbf{MIMIC}}
KNN and Gaussian Copula PIs yield the best overall performance, achieving the lowest CWFDC among all methods. Examining the components of CWFDC, we find that both methods achieve the best coverage -- surpassing even RUE with CP. % This aligns with qualitative assessments of the PIs (Figure~\ref{fig:pi_lineplot_comparison_mimic}): from time 500 to 1000, the spikes in actual (blue) breaths per minute (RESP) fall within the prediction intervals for both Gaussian Copula and KNN, unlike in RUE CP, illustrating their superior coverage.

These observations highlight the benefit of leveraging feature-wise rather than aggregated uncertainty, as is done in CP. Aggregated uncertainty reduces the likelihood of detecting anomalies in individual features -- for example, peaks in heart rate -- which can be hidden when averaged across all features. As a tradeoff for improved coverage, both methods produce slightly wider PIs (wider intervals increase coverage).

\paragraph{\textbf{PhysioNet}}
Comparing the PIs based on CWFDC, the Gaussian Copula PI emerges as the best performer, followed closely by RUE with CP. When examining each component of CWFDC, we again observe Gaussian Copula’s superior coverage compared to RUE CP. This is also evident in the qualitative evaluations shown in Figure~\ref{fig:pi_lineplot_comparison_physionet}: RUE CP’s prediction interval failed to capture the drop in Systolic ABP at $t=46$, unlike the other multidimensional RUE-based PIs. This further highlights the advantage of leveraging feature-wise uncertainty. In addition to its improved coverage, Gaussian Copula also achieves the smallest interval width. By contrast, KNN PI exhibited poorer coverage on this dataset, a deviation from the trend observed on MIMIC.

\begin{comment}
\paragraph{\textbf{Other RUE-derived PIs}}
In both datasets, the conditional Gaussian PI performed poorly across most metrics on MIMIC and showed weak coverage on PhysioNet. Its inferior performance compared to the Gaussian Copula PI suggests that the marginal distributions of prediction and reconstruction errors deviate from the Gaussian assumption. We validate this by testing each variable’s distribution using the Kolmogorov–Smirnov test for goodness of fit against a normal distribution. Combining the resulting p-values using Fisher’s method yields an overall p-value close to 0, indicating strong evidence against normality. By computing the correlation between the weights and the prediction error distances, we observe weak negative correlations for both datasets (–0.183 for MIMIC and –0.117 for PhysioNet). This suggests that reconstruction error distances are not effective for weighting in PI generation, which explains the poor performance of the Weighted PI method.
\end{comment}

\paragraph{\textbf{Gaussian Copula vs. KNN}}
The choice between Gaussian Copula and KNN for PI generation depends on three key dataset attributes: (1) the size of the calibration set, (2) the distribution of errors, and (3) the temporal resolution of the data. In our experiments, we found that when the calibration set is small (as in PhysioNet), KNN struggles to estimate PIs accurately, resulting in wider intervals. Gaussian Copula tends to perform better when the reconstruction and prediction errors are closer to Gaussian, as indicated by the Henze–Zirkler (HZ) test statistic. For example, the HZ statistic is lower for PhysioNet (1.5) than for MIMIC (2.4), suggesting that the errors in PhysioNet are more Gaussian -- consistent with our observation that Gaussian Copula performs better on PhysioNet than on MIMIC. Furthermore, Gaussian Copula performed the worst on MIMIC at t+2, where the HZ statistic was particularly high (3.3). Temporal resolution also plays a role: minute-level data in MIMIC captures local fluctuations, allowing KNN to find similar short-term patterns and produce intervals with better coverage, whereas the hour-level data in PhysioNet is smoother and more Gaussian-like, favoring the parametric approach of Gaussian Copula.

\section{Discussion and Conclusion}\label{sec:conclusion}
This paper introduces two methods -- Gaussian Copula and KNN -- for deriving prediction intervals (PIs) from the multidimensional Reconstruction Uncertainty Estimate. These methods focus on estimating the conditional distribution of the PI based on the input's uncertainty. The Gaussian Copula models this distribution directly by fitting a multivariate Gaussian Copula to reconstruction and prediction errors. The KNN method estimates the PI empirically using prediction errors of instances with similar reconstruction errors. Experiments on the MIMIC and PhysioNet datasets show that the Gaussian Copula method performs consistently well across both datasets, achieving high coverage with competitive widths. The KNN method outperforms on MIMIC, where the higher temporal resolution and larger calibration set favour its local estimation approach.

As all proposed PIs support multidimensional uncertainty estimates, future studies could explore using them to ensemble uncertainty estimates and capture all sources of prediction uncertainty (aleatoric, epistemic, and distributional). Additionally, while RUE-based PIs show promise, like CP, their accuracy depends on validation set size, with smaller sets affecting reliability. Future work could develop direct PI generation methods, such as training an uncertainty calibration network with a loss function optimizing coverage and interval width, reducing reliance on validation sets.
% For instance, concatenating MCD, Deep Evidential Regression and RUE and applying the Gaussian Copula method to estimate the PI conditioned from all three uncertainty estimates.

\bibliographystyle{IEEEtran}
\bibliography{citations}

\appendices

\section{Dataset Pre-Processing}
\label{sec:preprocessing}

Preprocessing steps for MIMIC dataset:
\begin{enumerate}
    \item Select patients with all six vital signs: [\enquote{ABPdias (mmHg)}, \enquote{ABPmean (mmHg)}, \enquote{ABPsys (mmHg)}, \enquote{HR (bpm)}, \enquote{RESP (bpm)}, \enquote{SpO2 (\%)}]. A total of 57 patients were selected.
    \item Randomly assign 70\%, 10\%, and 20\% of patients to the training, validation, and test sets, respectively. %The prediction model will be trained on the training set, its parameters tuned on the validation set, and its performance evaluated on the test set. 
    \item Remove all feature values less than 0 or blood pressure values greater than 250.
    \item Z-normalize the signals using the mean and standard deviation derived from the training set.
    \item To reduce the signal frequency, downsample the signal to one sample per minute by calculating the mean and standard deviation of signals within each minute.
    \item Drop rows with missing values.
\end{enumerate}

Preprocessing steps for PhysioNet Challenge dataset:
\begin{enumerate}
    \item Rows with missing values were dropped.
    \item Rows with anomalous values were dropped, such as when Diastolic, Mean, or Systolic ABP was 0 or below, or Urine Output exceeded 1,000 mL.
    \item Data were discretized into 1-hour intervals, as this was the most common sampling frequency.
    \item Set A was split into training and validation subsets on a per-patient basis; 10\% of patients were randomly assigned to the validation set.
    \item All signals were min-max normalized.
\end{enumerate}

\enquote{ABPdias (mmHg)}, \enquote{ABPmean (mmHg)}, \enquote{ABPsys (mmHg)} signals correspond to the diastolic, mean, and systolic arterial blood pressure of the patient, while \enquote{HR (bpm)}, \enquote{RESP (bpm)}, \enquote{SpO2 (\%)} correspond to periodic measurements of the heart rate, respiration rate, and oxygen saturation of the patient.

\section{Implementation Details}
To expedite the search process in step \ref{itm:knn} of KNN PI, we employ a K-Dimensional Tree \cite{friedman1977algorithm} fitted on validation set reconstruction errors before inference, and queried with the reconstruction errors of $\x$ during inference. We applied RUE, MCD, and IN to an MLP prediction model, using an MLP decoder for RUE. For RUE, encoder and decoder width/depth were fine-tuned via grid search, with validation loss and correlation with loss as their respective metrics. Regularisation and early stopping ($\text{patience}=20$) mitigated overfitting. Hyperparameters of other baselines were tuned similarly via grid search using validation performance. All experiments were repeated five times, reporting mean and standard deviation of all metrics. For implementation details, hyperparameter settings, and ablation studies on the KNN parameter $k$, refer to the code repository: \url{https://github.com/lr98769/rue_prediction_interval}.

\section{Evaluating Time Series Prediction Performance}
\begin{table}[ht]
    \centering
    \caption{Comparison of Mean Squared Error of Models Across All Time Horizons. In each column, we \textbf{bold} the values for the best-performing model and \underline{underline} the values for the second-best-performing model.}
     \scriptsize
\begin{tabular}{ccc}
\toprule
    \textbf{Model} & \textbf{MIMIC} & \textbf{PhysioNet}\\
\midrule
    RUE & \underline{0.1008 $\pm$ 0.000} & \underline{0.0024 $\pm$ 0.000}\\
\midrule
    MCD & 0.1017 $\pm$ 0.000 & \underline{0.0024 $\pm$ 0.000} \\
    GPR & \textbf{0.0962 $\pm$ 0.000} & \textbf{0.0023 $\pm$ 0.000}  \\
    IN & \underline{0.1008 $\pm$ 0.000} & \underline{0.0024 $\pm$ 0.000}\\
    BNN & 0.1028 $\pm$ 0.000 & 0.0053 $\pm$ 0.000 \\
    DER & 0.1010 $\pm$ 0.000 & 0.0026 $\pm$ 0.000 \\
\bottomrule
\end{tabular}
    \label{tab:pred_perf}
\end{table}
Table~\ref{tab:pred_perf} presents the average forecasting performance of various uncertainty estimation models across three time horizons. For the MIMIC dataset, SGPR achieved the best prediction performance, followed by IN and RUE (or multilayer perceptron (MLP)), as evidenced by their low MSE values across all horizons. On the PhysioNet dataset, RUE, MCD, SGPR and IN had comparable performances. Across both datasets, BNN and DER exhibited the weakest performance. These results demonstrate that the choice of uncertainty estimation method can significantly influence prediction performance.

\section{Evaluating Uncertainty Estimation Performance}
We evaluate uncertainty estimates using three metrics, with prediction loss measured by mean absolute error (MAE).

\paragraph{\textbf{Correlation}} The correlation between the uncertainty estimate and prediction error \cite{miTrainingFreeUncertaintyEstimation2022}, where a strong positive correlation indicates a reliable estimate.

\paragraph{\textbf{AURC}} The area under the risk-coverage curve \cite{ding2020revisiting}, which plots prediction loss (risk) against the proportion of confident predictions (coverage). Lower AURC indicates a more selective uncertainty estimate.

\paragraph{\textbf{$\boldsymbol{\sigma}$-Risk Score}} Measures prediction error for confident predictions (normalized uncertainty $\leq \sigma \in [0.1,0.2]$), reflecting robustness to false negatives (incorrect predictions with high confidence). Lower values indicate greater resilience. To mitigate the impacts of outliers, uncertainty is normalized by (1) excluding values beyond 1.5 IQR above Q3 and (2) applying min–max normalization on the remaining range.

\begin{table}[t]
    \centering
    \caption{Comparison of Uncertainty Estimation Performance on MIMIC and PhysioNet Across Time Horizons. For each metric, we \textbf{bold} values for the best-performing model and \underline{underline} values for the second-best-performing model.}
    \setlength{\tabcolsep}{2pt}
% \begin{adjustbox}{width = 0.8\textwidth, center}
    \scriptsize
    \begin{tabular}{P{0.8cm}cccccc}
\toprule
    \textbf{Data} & \textbf{t+k} & \textbf{Model} & \textbf{Correlation ($\uparrow$)} & \textbf{AURC ($\downarrow$)} & \textbf{$\boldsymbol{\sigma}$ = 0.1 ($\downarrow$)} & \textbf{$\boldsymbol{\sigma}$ = 0.2 ($\downarrow$)} \\
\midrule
    \multirow[c]{18}{*}{MIMIC} & \multirow[c]{6}{*}{t+1} & RUE & \textbf{0.431 $\pm$ 0.038} & 0.095 $\pm$ 0.003 & \underline{0.062 $\pm$ 0.007} & 0.079 $\pm$ 0.004 \\
    & & MCD & 0.192 $\pm$ 0.028 & 0.125 $\pm$ 0.001 & 0.103 $\pm$ 0.006 & 0.121 $\pm$ 0.002 \\
    & & GPR & 0.205 $\pm$ 0.002 & \underline{0.084 $\pm$ 0.000} & 0.070 $\pm$ 0.003 & \underline{0.076 $\pm$ 0.000}\\
    & & IN & 0.086 $\pm$ 0.025 & 0.128 $\pm$ 0.002 & 0.124 $\pm$ 0.009 & 0.126 $\pm$ 0.005 \\
    &  & BNN & \underline{0.285 $\pm$ 0.017} & 0.116 $\pm$ 0.001 & 0.081 $\pm$ 0.004 & 0.104 $\pm$ 0.004 \\
    &  & DER & 0.242 $\pm$ 0.035 & \textbf{0.074 $\pm$ 0.002} & \textbf{0.057 $\pm$ 0.003} & \textbf{0.063 $\pm$ 0.002}\\
\cmidrule{2-7}
    &  \multirow[c]{6}{*}{t+2} & RUE & \textbf{0.384 $\pm$ 0.019} & \textbf{0.109 $\pm$ 0.005} & \textbf{0.088 $\pm$ 0.008} & \textbf{0.096 $\pm$ 0.007}\\
    &  & MCD & 0.132 $\pm$ 0.010 & 0.151 $\pm$ 0.001 & 0.113 $\pm$ 0.005 & 0.142 $\pm$ 0.008 \\
    &  & GPR & 0.170 $\pm$ 0.003 & \underline{0.112 $\pm$ 0.000} & 0.102 $\pm$ 0.001 & 0.103 $\pm$ 0.000 \\
    &  & IN & -0.002 $\pm$ 0.007 & 0.156 $\pm$ 0.002 & 0.173 $\pm$ 0.033 & 0.163 $\pm$ 0.006 \\
    &  & BNN & \underline{0.203 $\pm$ 0.009} & 0.145 $\pm$ 0.003 & 0.108 $\pm$ 0.011 & 0.132 $\pm$ 0.004 \\
    &  & DER & 0.160 $\pm$ 0.066 & \underline{0.112 $\pm$ 0.002} & \underline{0.093 $\pm$ 0.003} & \underline{0.100 $\pm$ 0.004}\\
\cmidrule{2-7}
    &  \multirow[c]{6}{*}{t+3} & RUE & \textbf{0.366 $\pm$ 0.012} & \underline{0.138 $\pm$ 0.006} & \textbf{0.101 $\pm$ 0.024} & \underline{0.125 $\pm$ 0.006}\\
    &  & MCD & 0.120 $\pm$ 0.023 & 0.172 $\pm$ 0.003 & 0.122 $\pm$ 0.005 & 0.165 $\pm$ 0.005 \\
    &  & GPR & 0.152 $\pm$ 0.003 & \textbf{0.132 $\pm$ 0.001} & 0.122 $\pm$ 0.002 & \textbf{0.121 $\pm$ 0.000}\\
    &  & IN & -0.001 $\pm$ 0.007 & 0.178 $\pm$ 0.005 & 0.160 $\pm$ 0.035 & 0.173 $\pm$ 0.010 \\
    &  & BNN & \underline{0.174 $\pm$ 0.012} & 0.167 $\pm$ 0.001 & 0.124 $\pm$ 0.006 & 0.157 $\pm$ 0.003 \\
    &  & DER & 0.120 $\pm$ 0.020 & 0.139 $\pm$ 0.004 & \underline{0.120 $\pm$ 0.002} & 0.132 $\pm$ 0.009 \\
\midrule
    \multirow[c]{18}{1cm}{\centering Physio Net} & \multirow[c]{6}{*}{t+1} & RUE & \textbf{0.282 $\pm$ 0.014} & \underline{0.025 $\pm$ 0.001} & \textbf{0.020 $\pm$ 0.002} & \textbf{0.022 $\pm$ 0.001} \\
    &  & MCD & 0.189 $\pm$ 0.012 & 0.026 $\pm$ 0.001 & 0.024 $\pm$ 0.001 & 0.025 $\pm$ 0.001 \\
    & & GPR & 0.174 $\pm$ 0.001 & \textbf{0.024 $\pm$ 0.000} & \underline{0.022 $\pm$ 0.000} & 0.024 $\pm$ 0.000 \\
    & & IN & 0.057 $\pm$ 0.012 & 0.028 $\pm$ 0.001 & 0.028 $\pm$ 0.003 & 0.027 $\pm$ 0.001 \\
    & & BNN & 0.122 $\pm$ 0.004 & 0.050 $\pm$ 0.000 & 0.049 $\pm$ 0.003 & 0.048 $\pm$ 0.001 \\
    & & DER & \underline{0.236 $\pm$ 0.013} & \underline{0.025 $\pm$ 0.001} & \underline{0.022 $\pm$ 0.001} & \underline{0.023 $\pm$ 0.001} \\
\cmidrule{2-7}
    & \multirow[c]{6}{*}{t+2} & RUE & \textbf{0.232 $\pm$ 0.007} & \underline{0.029 $\pm$ 0.000} & \textbf{0.025 $\pm$ 0.001} & \textbf{0.026 $\pm$ 0.000} \\
    &  & MCD & 0.149 $\pm$ 0.007 & 0.030 $\pm$ 0.000 & 0.029 $\pm$ 0.001 & 0.028 $\pm$ 0.001 \\
    &  & GPR & 0.148 $\pm$ 0.000 & \textbf{0.028 $\pm$ 0.000} & \underline{0.026 $\pm$ 0.000} & 0.028 $\pm$ 0.000 \\
    &  & IN & 0.035 $\pm$ 0.006 & 0.032 $\pm$ 0.000 & 0.042 $\pm$ 0.009 & 0.031 $\pm$ 0.002 \\
    &  & BNN & 0.111 $\pm$ 0.005 & 0.051 $\pm$ 0.000 & 0.049 $\pm$ 0.003 & 0.048 $\pm$ 0.001 \\
    &  & DER & \underline{0.173 $\pm$ 0.007} & \underline{0.029 $\pm$ 0.001} & 0.029 $\pm$ 0.002 & \underline{0.027 $\pm$ 0.001} \\
\cmidrule{2-7}
    & \multirow[c]{6}{*}{t+3} & RUE & \textbf{0.231 $\pm$ 0.009} & \underline{0.031 $\pm$ 0.001} & \textbf{0.028 $\pm$ 0.001} & \textbf{0.028 $\pm$ 0.001} \\
    &  & MCD & 0.136 $\pm$ 0.015 & 0.033 $\pm$ 0.000 & 0.032 $\pm$ 0.001 & 0.031 $\pm$ 0.001 \\
    &  & GPR & 0.147 $\pm$ 0.000 & \textbf{0.030 $\pm$ 0.000} & \underline{0.029 $\pm$ 0.000} & \underline{0.030 $\pm$ 0.000} \\
    &  & IN & 0.030 $\pm$ 0.007 & 0.034 $\pm$ 0.000 & 0.038 $\pm$ 0.005 & 0.033 $\pm$ 0.002 \\
    &  & BNN & 0.105 $\pm$ 0.007 & 0.051 $\pm$ 0.000 & 0.049 $\pm$ 0.003 & 0.049 $\pm$ 0.001 \\
    &  & DER & \underline{0.186 $\pm$ 0.019} & 0.032 $\pm$ 0.001 & 0.031 $\pm$ 0.003 & 0.031 $\pm$ 0.001 \\
\bottomrule
    \end{tabular}
% \end{adjustbox}
    %\vspace{-10pt}
    \label{tab:ue_perf}
\end{table}

Table~\ref{tab:ue_perf} compares uncertainty estimate performance across datasets and horizons. Reliability generally decreases with longer horizons, as shown by the declining correlation with prediction error from $t+1$ to $t+3$, suggesting existing estimates struggle to capture uncertainty at extended horizons.

\textbf{Correlation:} RUE is the most reliable estimate on both MIMIC and PhysioNet, followed by BNN on MIMIC and DER on PhysioNet. % This suggests distributional uncertainty dominates both datasets, with higher epistemic and aleatoric uncertainty in MIMIC and PhysioNet, respectively.

\textbf{AURC:} SGPR achieves the lowest AURC, followed closely by RUE, particularly beyond $t+1$. This highlights RUE's potential for selective prediction (Table~\ref{tab:pred_perf}). % , yielding low selective prediction loss even on models with lower overall performance

\textbf{$\sigma$-Risk Scores:} RUE is the most robust to false negatives, showing the smallest or second-smallest $\sigma$-risk across datasets and horizons, followed by DER and SGPR, indicating its confident predictions are less likely to be incorrect.

\end{document}